\documentclass[10pt,twocolumn,letterpaper]{article}

\usepackage{cvpr}
\usepackage{times}
\usepackage{epsfig}
\usepackage{graphicx}
\usepackage{amsmath}
\usepackage{amssymb}
\usepackage{bbding}
\usepackage[ruled]{algorithm2e}
\usepackage[breaklinks=true,bookmarks=false]{hyperref}

\cvprfinalcopy
\def\cvprPaperID{483}

\ifcvprfinal\fi

\begin{document}

\title{Spatiotemporal Pyramid Network for Video Action Recognition\thanks{Presented at CVPR 2017. Contact author: Mingsheng Long.}}

\author{
    Yunbo Wang$^\dag$, Mingsheng Long$^\dag$, Jianmin Wang$^\dag$, and Philip S. Yu$^{\dag\ddag}$\\
    $^\dag$KLiss, MOE; TNList; NEL-BDSS; School of Software, Tsinghua University, China\\
    $^\ddag$University of Illinois at Chicago, IL, USA\\
    {\tt\small wangyb15@mails.tsinghua.edu.cn, \{mingsheng,jimwang\}@tsinghua.edu.cn, psyu@uic.edu}
}

\maketitle

\thispagestyle{empty}

\begin{abstract}
   Two-stream convolutional networks have shown strong performance in video action recognition tasks. The key idea is to learn spatiotemporal features by fusing convolutional networks spatially and temporally. However, it remains unclear how to model the correlations between the spatial and temporal structures at multiple abstraction levels. First, the spatial stream tends to fail if two videos share similar backgrounds. Second, the temporal stream may be fooled if two actions resemble in short snippets, though appear to be distinct in the long term. We propose a novel spatiotemporal pyramid network to fuse the spatial and temporal features in a pyramid structure such that they can reinforce each other. From the architecture perspective, our network constitutes hierarchical fusion strategies which can be trained as a whole using a unified spatiotemporal loss. A series of ablation experiments support the importance of each fusion strategy. From the technical perspective, we introduce the spatiotemporal compact bilinear operator into video analysis tasks. This operator enables efficient training of bilinear fusion operations which can capture full interactions between the spatial and temporal features. Our final network achieves state-of-the-art results on standard video datasets.
\end{abstract}

\vspace{-30pt}

\section{Introduction}

Learning a good video representation is the foundation of many computer vision tasks, such as action recognition and video captioning. It goes beyond image analysis and depends on a joint modeling of both spatial and temporal cues. Many existing methods~\cite{Karpathy14,Ji13,Tran15,sun2015human} are dedicated to this modeling by taking advantages of Convolutional Neural Networks (CNN)~\cite{lecun1998gradient,krizhevsky2012imagenet,simonyan2015very,szegedy2015going}. However, these CNN-based methods have not shown an overwhelming performance over other approaches~\cite{wang2013action,wang2015action} using Fisher Vector~\cite{perronnin2010improving}, HOF~\cite{laptev2008learning}, and dense trajectories~\cite{wang2013action}. One reason is that these CNN frameworks are not specifically designed for videos and cannot fully exploit spatiotemporal features. 

\begin{figure}[t]
\begin{center}
   \includegraphics[width=0.45\textwidth]{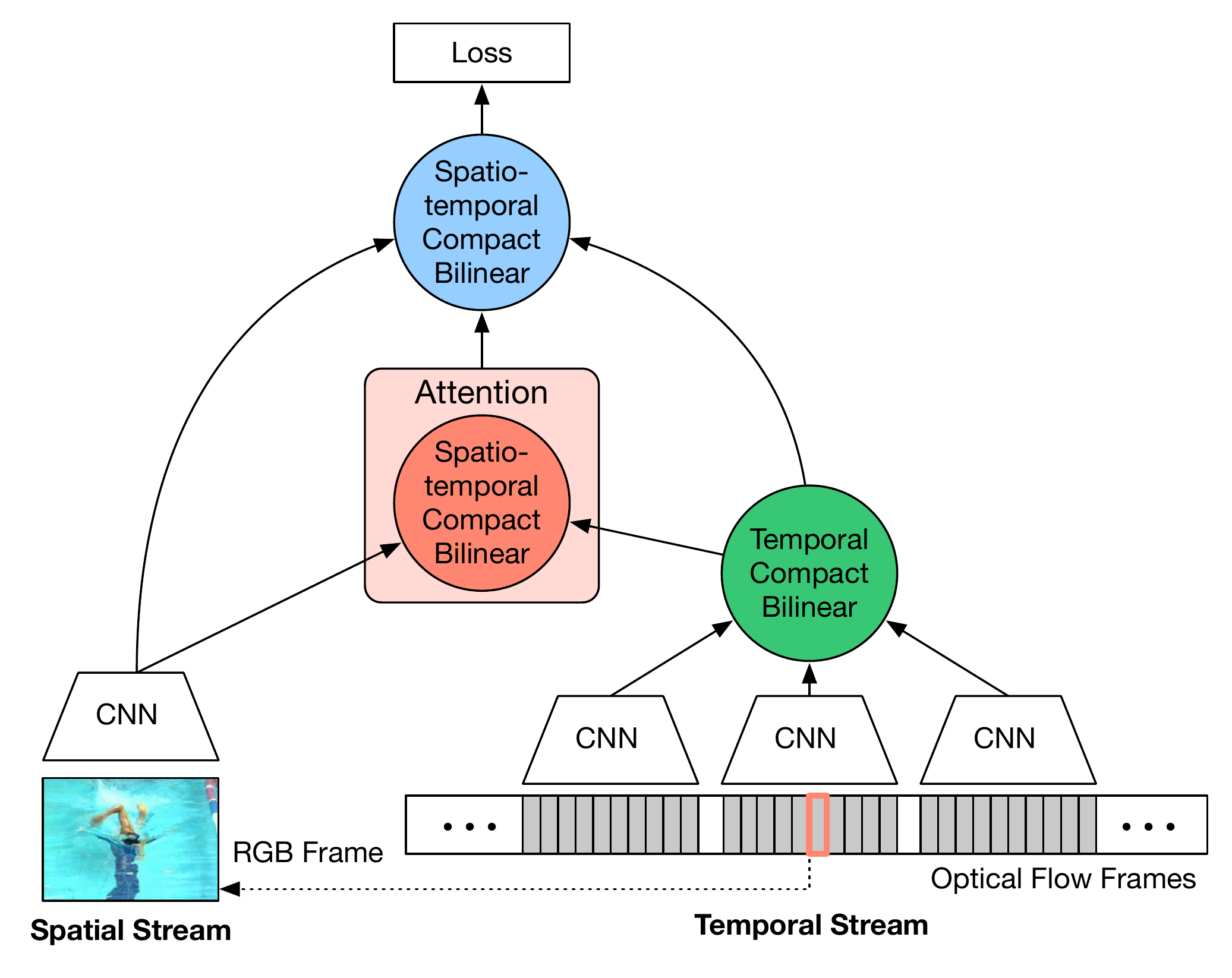}
\end{center}
\vspace{-10pt}
   \caption{An overview of our spatiotemporal pyramid network, which constitutes a multi-level fusion pyramid of spatial features, long-term temporal features and spatiotemporal attended features.}
\label{fig:overview}
\vspace{-15pt}
\end{figure}

In addition to capturing the appearance information using standard CNN stream, several recent approaches try out using optical flow data in a second CNN stream to capture the motion information~\cite{Simonyan14,Ng15,Feichtenhofer16,zhang2016real}. However, when taking a closer look at these models, we observe that for most misclassification cases, there is usually one stream failing, while the other remaining correct. Hence, simply averaging the outputs of the classifier layers is not enough. Instead, we hope to make the spatial and temporal cues facilitate each other. This paper presents a novel end-to-end spatiotemporal pyramid architecture, as shown in Figures~\ref{fig:overview}, which can on one hand boost the accuracy of individual stream and on the other hand exploits spatiotemporal cues jointly. 

From the temporal perspective, since the original optical flow stream only receives 10 consecutive optical flow frames, it may be fooled if two actions resemble in such a short snippet, though distinguish in the long term. A typical example is Pull-ups and RopeClimbing in UCF101: men in these two categories could be moving in the same single direction in short snippets. But if we enlarge the time range, we can easily find that the man in Pull-ups is actually moving up and down, while the one in RopeClimbing is moving straight upwards. To learn more global video features, we use multi-path temporal subnetworks to sample optical flow frames in a longer sequence, and explore several fusion strategies to combine the temporal information effectively.

From the spatial perspective, the original spatial stream is easily fooled when the backgrounds of two videos are extremely similar. For instance, it cannot tell FrontCrawl and BreastStroke apart, because for both categories, the swimming pool turns out to be the strongest feature. However, the optical flow network can tell these actions quite clearly and outperform the spatial stream by at least 5\% on UCF101. Motivated by this, we introduce a spatiotemporal attention module to extract significant locations on feature maps of the spatial network. In this process, the temporal features are exploited as a guidance, which informs the spatial stream where the motion of interest happens. 

Besides improving the effectiveness of individual streams, we explore methods to fuse the spatiotemporal features, which enable a joint optimization of the whole architecture using a unified spatiotemporal loss function. We bring in the compact bilinear fusion strategy, which captures full interactions across spatial and temporal features, while significantly reduces the number of parameters of traditional bilinear fusion methods from millions to just several thousands. Our experiment results demonstrate that compact bilinear approximately models the correlations between each single element of the spatial and temporal features, thus yields better performance over its alternatives, such as concatenation and element-wise sum studied in \cite{Feichtenhofer16}.

Our contributions can be summarized as follows. \textbf{(a)} We propose a novel deep learning architecture to address the problems we observe in video representation learning. \textbf{(b)} We introduce the compact bilinear and spatiotemporal attention methods into video-related tasks and validate their feasibility in practice. \textbf{(c)} We evaluate our approaches on standard video datasets UCF101 and HMDB51 and report significant improvement over the previous state-of-the-art.

\section{Related Work}

Motivated by the impressive performance of deep learning on image-related tasks, several recent works try to design effective CNN-based architectures for video recognition that jointly model spatial and temporal cues. Before the blossom of CNN, Ji et al.~\cite{Ji13} first exploit consecutive video frames as inputs and extend the convolutional filters into temporal domain. Karparthy et al.~\cite{Karpathy14} compare multiple CNN connectivity methods in time, including late fusion, early fusion and slow fusion. But these approaches cannot fully utilize motion information, and only yield a modest improvement over operating on single frames. Tran et al.~\cite{Tran15} train a deeper CNN model called C3D on Sports-1M. C3D is basically a 3D version of VGGnet~\cite{simonyan2015very}, containing 3D convolution filters and 3D pooling layers operating over space and time simultaneously. Noticing that stacked RGB frames cannot fully exploit temporal cues, Simonyan et al.~\cite{Simonyan14} train a second stream of CNN on optical flow frames and propose a two-stream ensemble network. Since the optical flow data brings in a significant performance gain, it has recently been employed into many other action recognition methods~\cite{cheron2015p,donahue2015long,srivastava2015unsupervised,venugopalan2015sequence,weinzaepfel2015learning,Ng15,sun2015human,Wangxiaolong16,Feichtenhofer16}. 

\begin{table}
\begin{center}
\begin{tabular}{|l|c|c|}
\hline
 & Ours & Two-Stream Fusion\\
\hline\hline
Arch. & Pyramid & Single Conv-Layer \\
\#Loss & 1 (End-to-End) & 2 (Average) \\
Spatial & Attention Pooling & Average Pooling \\
Temporal & Compact Bilinear & 3D Max Pooling \\
\hline
\end{tabular}
\end{center}
\vspace{-5pt}
\caption{Differences between our method and state-of-the-art~\cite{Feichtenhofer16}.}
\label{tab:diff}
\vspace{-15pt}
\end{table}

However, the original two-stream method \cite{Simonyan14} has two main drawbacks: First, it only incorporate 10 consecutive optical flow frames, so that it cannot capture long-term temporal cues. Second, it cannot learn the subtle spatiotemporal relationships. The spatial (RGB frames) and temporal (optical flow) streams are trained separately, and the final predictions are obtained by averaging the outputs of two classifiers. To mitigate these issues, Ng et al.~\cite{Ng15} investigate several pooling methods as well as the Long Short-Term Memory (LSTM)~\cite{Ng15} to fuse features across a longer video sequence. Wang et al.~\cite{Wang16} model long-term temporal structures by proposing a segmental network architecture with sparse sampling. Feichtenhofer et al.~\cite{Feichtenhofer16} study multiple ways of combining networks both spatially and temporally. They propose a spatiotemporal fusion method and claim that the two-stream networks should be fused at the last convolutional layer. Table~\ref{tab:diff} lists the main differences between our work and~\cite{Feichtenhofer16}. First and foremost, we propose a multi-layer pyramid fusion architecture, replacing a 3D convolutional layer and a pooling layer in~\cite{Feichtenhofer16}, to combine the spatial and temporal features at different abstraction levels. For individual streams, we upgrade the spatial subnetwork by replacing the original average pooling with a spatiotemporal attention module. This method makes the network concentrate on significant regions on static frames with the help of motion cues. Moreover, we introduce the compact bilinear operator for fusing multi-path optical flow features temporally. Finally, in training strategy of~\cite{Feichtenhofer16}, two losses are used in their objective function and the final prediction is obtained by averaging the outputs of two streams. In contrast, our fusion network is trained end-to-end with one single spatiotemporal loss function. Thus, all streams are optimized as a whole, resulting in an improved result.

\section{Spatiotemporal Pyramid Network}
The spatiotemporal pyramid network supports long-term temporal fusion and a visual attention mechanism. Also, we propose a new spatiotemporal compact bilinear operator to enable a unified modeling of various fusion strategies.

\subsection{Spatiotemporal Compact Bilinear Fusion}

The fusion of spatial and temporal features in compact representations proves to be the key to learning high-quality spatiotemporal features for video recognition. A good fusion strategy should maximally preserve the spatial and temporal information while maximize their interaction. Typical fusion methods including element-wise sum, concatenation, and bilinear fusion have been extensively evaluated in the convolutional two-stream fusion framework \cite{Feichtenhofer16}. However, element-wise sum and concatenation do not capture the interactions across the spatial and temporal features, so they may suffer form substantial information loss. Bilinear fusion allows all spatial and temporal features in different dimensions to interact with each other in a multiplicative way. Since our spatiotemporal pyramid constitutes spatial features, temporal features, and their hierarchy, the bilinear fusion is the only appropriate strategy for our approach.

Specifically, denote by ${x}$ and ${y}$ the spatial and temporal feature vectors respectively, the bilinear fusion is defined as ${z} = {\rm{vec}}({x} \otimes {y})$, where $\otimes$ denotes the outer product ${x}{y}^{\sf T}$, and ${\rm{vec}}$ denotes the vectorization of a vector. Bilinear fusion leads to high dimensional representations with million of parameters, which will make network training infeasible.

To circumvent the curse of dimensionality, we propose a Spatiotemporal Compact Bilinear (STCB) operator to enable various fusion strategies. We transform the outer product to a lower-dimensional space which avoids computing the outer product directly. As suggested by the compact bilinear pooling method \cite{gao2016compact}, for a single modality, we adopt the Count Sketch projection function $\Phi$ \cite{Charikar2002}, which projects a vector $v \in \mathbb{R}^p$ to $v' \in \mathbb{R}^{d}$. We initialize two vectors $s \in \{-1, 1\}^p$ and $h \in \{1,\dots,d\}^p$, where $s$ contains either $1$ or $-1$ for each index, and $h$ maps each index $j$ in the input $v$ to an index $k$ in the output $v'$. Both $s$ and $h$ are initialized randomly from a uniform distribution and remain constant for future invocations of Count Sketch. $v'$ is initialized as a zero vector. For every entry $v(j)$ its destination index $k = h(j)$ is looked up by $h$, and $s(j) \cdot v(j)$ is added to $v'(k)$. See Algorithm \ref{algorithm:STCB} for the details, where $m$ is the number of feature pathways for compact bilinear fusion.

This procedure enables projecting the outer product of spatial and/or temporal features into a lower-dimensional space, which significantly reduces the number of parameters from millions to several thousands. To avoid explicitly computing the outer product, \cite{Pham2013} reveals that the Count Sketch of the outer product of two vectors can be expressed as convolution of both Count Sketches: $\psi \left( {x \otimes y,h,s} \right) = \psi \left( {x,h,s} \right) * \psi \left( {y,h,s} \right)$, where $\ast$ is the convolution operator. Fortunately, the convolution theorem states that convolution in the time domain is equivalent to element-wise product in the frequency domain. Thus the convolution $x \ast y$ can be rewritten as ${\text{FF}}{{\text{T}}^{ - 1}}\left( {{\text{FFT}}\left( x \right) \odot {\text{FFT}}\left( y \right)} \right)$, where $\odot$ refers to element-wise product. These ideas are summarized in Algorithm \ref{algorithm:STCB}, which is based on the Tensor Sketch algorithm \cite{Pham2013}. We invoke the algorithm with $m$ pathways of spatial and/or temporal features that need to be fused, which enables spatiotemporal fusion into compact representations.

\begin{algorithm}[h]
    \DontPrintSemicolon
    \LinesNumbered
    \KwIn{Spatial and/or temporal features $\{v_i \in \mathbb{R}^{p_i}\}_{i=1}^m$}
    \KwOut{Fused features $\Phi(\{v_i\}_{i=1}^m) \in \mathbb{R}^d$}
    \For{$i \leftarrow 1$ \KwTo $m$}{
      \If{$h_i,s_i$ \emph{not initialized}}{
         \For{$j \leftarrow 1$ \KwTo $p_i$}{
            sample $h_i(j)$ from $\{1,\ldots,d\}$\;
            sample $s_i(j)$ from $\{-1,1\}$
         }
      }
      $v'_i = [0,\ldots,0]$\;
      \For{$j \leftarrow 1$ \KwTo $p_i$}{
         $v'_i\left( {h_i\left( j \right)} \right) = v'_i\left( {h_i\left( j \right)} \right) + s_i\left( j \right) \cdot {v_i}\left( j \right)$\;
      }
    }
    $\Phi \left( {\left\{ {{v_i}} \right\}_{i = 1}^m} \right) = {\text{FF}}{{\text{T}}^{ - 1}}\left( { \bigodot _{i = 1}^m{\text{FFT}}\left( {{{v'_i}}} \right)} \right)$\;
   \caption{STCB: Spatiotemporal compact bilinear}
   \label{algorithm:STCB}
\end{algorithm}

\begin{figure*}
\begin{center}
\includegraphics[width=1.0\textwidth]{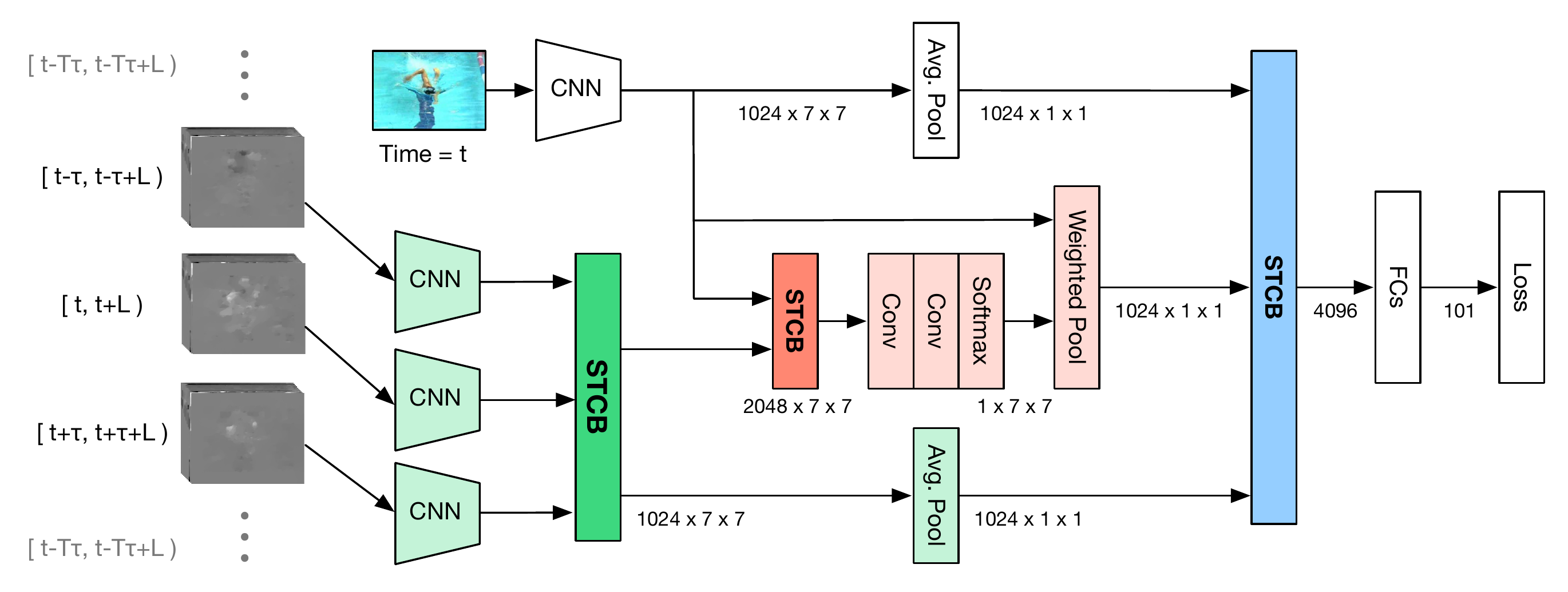}
\end{center}
\vspace{-5pt}
   \caption{The spatiotemporal pyramid network learns spatiotemporal features at multiple abstraction levels, which can be trained end-to-end as a whole. Optical flow features are first combined across time by a compact bilinear layer. The resulting features then run through the second compact bilinear layer and a spatiotemporal attention module, determining the salient regions of activities. The final video representations are obtained by fusing features from the spatial stream, the temporal stream and the attention stream.}
\label{fig:Arch}
\vspace{-5pt}
\end{figure*}

\vspace{-10pt}
\subsection{Temporal Fusion}

The original temporal stream takes 10 consecutive optical flow frames as inputs, thus it may make mistakes if two actions look similar in such short snippets, though differentiate in the long term. Therefore, we hypothesize that learning a more global representation would offer more accurate motion cues for the upper levels of the pyramid network. Specifically, we enlarge the input video chunks by using multiple CNNs with shared network parameters to sample the optical flow frames at an interval of $\tau$. Each chunk covers the previous $L/2$ and the next $L/2$ frames as inputs. For both training and testing, $L$ is fixed to 10, and $\tau$ is randomly selected from 1 to 10, in order to model variable lengths of videos with a fixed number of neurons. For the fusion method, we exploit STCB and make it support a scalable number of input feature maps. We show that STCB is effective not only for spatiotemporal fusion, but also for temporal combination. 

Comparing our method with \cite{Feichtenhofer16,Wang16}, all these three methods aim to broaden the input fields across the time domain. Multiple snippets are fused by 3D pooling in \cite{Feichtenhofer16} and by compact bilinear in our work. Another difference between our method and \cite{Feichtenhofer16} is that their temporal fusion includes fusing the features of multiple RGB frames as well, while we only combine optical flow representations. The reason is that modeling multiple RGB frames with another loss function would cause additional overfitting issue in training and obtain misleading results in testing. More importantly, in \cite{Feichtenhofer16,Wang16}, the resulting temporal features are directly fed into classifiers. In this paper, they are used as input to the next fusion stage (attention) in our architecture. We observe that compact bilinear fusion can preserve the temporal cues to supervise the spatiotemporal attention module.

\subsection{Spatiotemporal Attention}

The second level of our spatiotemporal fusion pyramid is a variant of the attention model, which is originally proposed in multi-modal tasks~\cite{yang2015stacked,xu2015show,lu2016hierarchical}. We adopt this idea and extensively apply it to the spatiotemporal scenario, by taking advantage of the motion information to locate salient regions on the image feature maps. We implement it on the last convolutional layers (i.e. inception5b in BN-Inception, res5c in ResNets and conv5 in VGGnet). For one thing, the representations of these layers show class-specific motion variations, while the lower layers capture finer-grained features of the image, such as edges, corners and textures. For another thing, we want the attention module to play a role as a more accurate weighted pooling operation, because we believe that the original average pooling cannot capture the salient regions corresponding to the activity information and may results in information loss. 

The spatiotemporal attention module reduces a $1048 \times 7 \times 7$ feature map in BN-Inception to a $1048 \times 1 \times 1$ feature vector. But unlike max pooling or average pooling, the attention pooling has a relatively sophisticated logic and complicated mechanism. More specifically, for each grid location on the image feature maps, we use STCB to merge the spatial feature vector with its temporal counterpart. The output spatiotemporal representations, implying the corresponding appearance and motion cues, serve as the supervision of the following attention layers. After that, two convolutional layers are stacked to produce the attention weights for the feature maps. The size of the first convolutional layer is $64 \times 7 \times 7$, while that of the second is $1 \times 7 \times 7$. At last, the resulting attention weights are normalized by a softmax layer, and then combined with the original spatial features by a weighted pooling layer. The spatiotemporal attention with STCB makes the spatial stream prone to be abstracted by the moving objects in the static RGB frames.

Though the attention mechanism has been explored in action recognition, 
our work differs from the others, such as ~\cite{Sharma16} in two folds: \textbf{(a)} our attention is generated by both spatial-stream and temporal-stream CNNs while the others' are generated by LSTMs, both to highlight the motion cues in the spatial representations; \textbf{(b)} we further use optical flow features as the temporal supervision to the attention module.

\subsection{Proposed Architecture}

All techniques mentioned above can be integrated under a pyramid framework. We design our architecture by injecting the proposed fusion layers between the convolutional and the fully connected layers. Under these circumstances, we only use the representations of the last convolutional layer, so that our approach can be extendible for almost all CNN architectures, including BN-Inception, ResNets and VGGnet. As a result, we can initialize our network with models that are well pre-trained on ImageNet~\cite{deng2009imagenet} before we fine-tune them on the relatively small video datasets.

We use the STCB technique three times. At the bottom of the pyramid, multiple optical flow representations across a longer video sequence are combined by the first STCB (green layers in Figure~\ref{fig:Arch}). By doing this, we obtain more global temporal features. These features are then fed into the next fusion level, the spatiotemporal attention subnetwork (red layers), where we use another STCB to fuse the spatial feature maps with the corresponding motion representations, and offer the attention cues of salient activities. At the top of the fusion pyramid, all the three previous outcomes are used: the original spatial and temporal features through average pooling, as well as the resulting attended features through the attention module. They are fused into a 4096-dimensional vector by a third STCB, and each of them captures significant information of multiple abstraction levels in the pyramid. Removing any feature pathway would result in a decrease in the overall performance. 

\section{Experiments}

This section is organized in accordance with the progress of our experiments. Initially, we describe the datasets and training details. Then, we explore the effects of applying different deep networks as the building blocks of our model, including VGGnet~\cite{simonyan2015very}, ResNets~\cite{He2015Deep} and BN-Inception~\cite{ioffe2015batch}. Next, we evaluate our spatiotemporal pyramid network and demonstrate its effectiveness by giving ablation results. Last but not least, we compare our method with the previous state-of-the-art and analyze its performance by giving typical examples of correct and incorrect predictions.

\subsection{Datasets and Implementations}

We train and evaluate our spatiotemporal pyramid network on two standard datasets. The UCF101 dataset~\cite{Soomro12} contains 13320 fully-annotated video snippets from 101 action categories. Each snippet lasts 3-10 seconds and consists of 100-300 frames on average. The HMDB51 dataset~\cite{Kuehne12} contains 6766 videos clips that covers 51 action categories. For both of them, we follow the provided evaluation protocol and adopt standard training/testing splits.

To verify the pure effectiveness of different pyramid fusion layers, we train a VGG-16 model, a BN-Inception model and a ResNet-50 model respectively on standard video datasets. Our models are trained following a multi-stage training strategy. We initialize the spatial and the temporal subnetworks with models pre-trained on ImageNet. Then we fine-tune each of them on the target video datasets and fill these parameters into our final pyramid network. We follow the cross modality fine-tuning strategy presented in \cite{Wang16}. After that, we train the entire network on UCF101 or HMDB51. Here we choose the mini-batch stochastic gradient descent algorithm and set the batch size to 32. Moreover, for VGG-16, we stack the two 4096-dimensional fully connected layers behind the last STCB layer. We set the base learning rate to 0.01 and decrease it by a factor of 10 every 10,000 iterations, and the training process stops at iteration 30,000. For ResNets, the base learning rate is initialized as 0.001, reduce by a factor of 10 every 10,000 iterations, and stopped at 20,000. To avoid overfitting, we randomly sample the temporal interval $\tau$ from 1 to 10. Also, we exploit several data augmentation techniques, such as scale jittering, horizontal flipping and image cropping. Details of these tricks are not in the scope of this paper. All experiments are implemented with Caffe~\cite{Jia14}.

\subsection{Base Architectures}

Deeper CNNs can often lead to better performance in image recognition tasks~\cite{cimpoi2015deep,lin2015bilinear,szegedy2015going,He2015Deep}, since they bring in great modeling capacity and are capable of learning discriminative representation from raw visual data. The state-of-the-art two-stream architecture~\cite{Feichtenhofer16} is based on VGG-16~\cite{simonyan2015very}, while Inception with Batch Normalization (BN-Inception)~\cite{ioffe2015batch} and Deep Residual Networks (ResNets)~\cite{He2015Deep} have shown remarkable performance in several challenging recognition tasks recently. In this work we further explore the feasibility of ResNets in video analysis tasks. We take into account a 50-layer ResNet for the sake of computational complexity, as well as a 152-layer ResNet for its compelling accuracy. All models are pre-trained on the ImageNet~\cite{deng2009imagenet} and fine-tuned on UCF101 and HMDB51. 

Table~\ref{tab:RES} compares the performance of VGGnet, BN-Inception and ResNets. Generally, as the number of convolutional layers grows, the RGB network benefits most. In contrast, the performance of the optical flow network decreases slightly. There are two reasons. First, the optical flow data yields a different distribution from RGB, which weakens the impact of fine-tuning. Second, due to the limited amount of training samples on UCF101, complex network structures are prone to over-fitting. BN-Inception turns out to be the top-performing base architecture.
\begin{table}[h]
\begin{center}
\begin{tabular}{|l|c|c|c|}
\hline
Model & Spatial & Temporal & Two-Stream~\cite{Simonyan14} \\
\hline\hline
VGG-16 & 80.5\% & 85.4\% & 88.9\% \\
ResNet-50 & 83.7\% & 84.9\% & 90.3\% \\
ResNet-152 & 84.3\% & 82.1\% & 89.8\% \\
BN-Inception & 84.5\% & 87.0\% & 91.7\% \\
\hline
\end{tabular}
\end{center}
\caption{Classification accuracy of the two-stream model~\cite{Simonyan14} with different base architectures on UCF101 (Split 1). All results are obtained by averaging the outputs of the Softmax layers as \cite{Simonyan14}.}
\label{tab:RES}
\vspace{-10pt}
\end{table}

\subsection{Spatiotemporal Compact Bilinear Fusion}

We explore several strategies for fusing spatial and temporal feature maps. All models but the VGGnet one follow the same architecture, that the fusion layer is put between the last convolutional layer (i.e. res5c for ResNets and inception5b for BN-Inception) and the final classifier. Our experiments show that such a late fusion architecture outperforms its alternatives in which the fusion layer is moved forward. It can be explained by that the last convolutional layer shows class-specific and highly informative features with significant motion variations.

As shown in Table~\ref{tab:STF}, spatiotemporal compact bilinear fusion results in the highest accuracy and improves the performance by around 1.5 points. It is a valuable observation. Before this, what we know is that compact bilinear pooling is effective for combining visual representations. But what we do not know is that the same merit happens between spatial and temporal data. Table~\ref{tab:STF} also reveals that the output dimension makes a difference on the performance of spatiotemporal compact bilinear fusion. As it grows, the correlations between the spatial and the temporal representations (both 1024-dimensional) can be captured more completely, thus the classification accuracy increases. It is a trade-off between compression and quality. But larger output dimension is not always good, since it makes the following fully connected layers hard to train. We observe that a 4096 output dimension is appropriate for both video datasets.

\begin{table}[h]
\begin{center}
\begin{tabular}{|l|c|}
\hline
Fusion Method & Accuracy \\
\hline\hline
Average & 91.7\% \\
Concatenation & 92.4\% \\
Element-wise Sum & 92.3\% \\
Compact Bilinear (d = 1024) & 92.4\% \\
Compact Bilinear (d = 2048) & 92.9\% \\
Compact Bilinear (d = 4096) & 93.2\% \\
Compact Bilinear (d = 8192) & 93.2\% \\
\hline
\end{tabular}
\end{center}
\caption{Accuracy of various fusion methods on UCF101 (Split 1).}
\label{tab:STF}
\vspace{-5pt}
\end{table}

\subsection{Temporal Fusion}

Table~\ref{tab:MTF} illustrates the impact of feeding temporal networks with longer sequences of optical flow data. Our model is implemented by making several copies of the individual network and combining them at the last convolutional layers. We have two observations here. First, among all these fusion strategies, spatiotemporal compact bilinear fusion presents the best performance. It is the first time that compact bilinear fusion is demonstrated effective for merging multi-path optical flow representations.

Second, these results explain why we design 3 subnetworks in out final architecture. The columns in Table~\ref{tab:MTF} denotes the number of pathways before the fusion layer. Among all these models, a 3-path network with spatiotemporal compact bilinear fusion outperforms the others. We shall not cut down or increase the number of subnetworks. On one hand, the performance of the 3-path model is 2.3 points higher than that of the single-path. On the other hand, more subnetworks do not mean better results, since in this situation the spatial and temporal features may not correlate well with a very long sequence of optical flow data. 

\begin{table}[h]
\begin{center}
\begin{tabular}{|l|c|c|c|}
\hline
Fusion Method & 1-path & 3-path & 5-path \\
\hline\hline
Concatenation & 87.0\% & 88.4\% & 88.5\% \\
Element-wise Sum & - & 87.9\% & 87.7\% \\
Compact Bilinear & - & 89.3\% & 89.2\% \\
\hline
\end{tabular}
\end{center}
\caption{A comparison of methods for merging multi-path temporal chunks. The columns represent the number of temporal chunks. All results are produced on UCF101 with optical flow data only.}
\label{tab:MTF}
\end{table}

\subsection{Spatiotemporal Attention}

Attention pooling can effectively improve the overall classification accuracy by guiding the spatial network to attend to significant locations. Our experiments demonstrate that it can help avoid classification errors especially resulting from similar or chaotic backgrounds in static video frames. As shown in Table~\ref{tab:SAP}, our best implementation boosts the performance of the spatial pathway by 2.1 points. 

Moreover, this set of experiments testify the value of compact bilinear fusion again. We initially intend to use temporal representations solely to generate attention weights. However, the result turns out to be a little lower than the original average pooling. We then try to merge temporal and spatial features in advance, while in this scenario the compact bilinear fusion performs surprisingly well. 

\begin{table}[h]
\begin{center}
\begin{tabular}{|l|c|}
\hline
Model & Spatial Accuracy \\
\hline\hline
Average Pooling & 84.5\% \\
Att. Pooling (Temporal Only) & 84.3\%\\
Att. Pooling (Concatenation) & 83.9\%\\
Att. Pooling (Element-Wise Sum) & 83.5\%\\
Att. Pooling (Compact Bilinear) & 86.6\%\\
\hline
\end{tabular}
\end{center}
\caption{The effect of applying attention pooling to the spatial network on UCF101. We feed the attention module with representations generated by various fusion methods.}
\label{tab:SAP}
\end{table}

\subsection{Ablation Results}

To testify the individual effect of fusion approaches we discuss above, we stack them one by one and test the overall performance. We set the baseline as the original two-stream CNNs that averages the outputs of the classifier layers. From Table~\ref{tab:ablation}, we observe that our spatiotemporal fusion method improves the average accuracy by 1.5 points. Furthermore, the proposed multi-path temporal fusion method results in another 0.4 points performance gain. At last, we apply spatiotemporal attention pooling and boost the final result to 94.2\%. To sum up, all methods that we propose prove to be effective for video action recognition. 

\begin{table}[h]
\begin{center}
\begin{tabular}{|l|c|c|c|c|}
\hline
Model & A & B & C & D \\
\hline\hline
ST Fusion & - & \checkmark & \checkmark & \checkmark \\
Multi-T Fusion & - & - & \checkmark & \checkmark\\
Attention & - & - & - & \checkmark \\
\hline
Accuracy & 91.7\% & 93.2\% & 93.6\% & 94.2\% \\
\hline
\end{tabular}
\end{center}
\caption{Ablation results on UCF101 (Split 1). ST Fusion denotes two-stream spatiotemporal compact bilinear fusion. Multi-T Fusion denotes multi-path temporal fusion. Model A stands for the original two-stream CNNs, while the others stacks the proposed approaches one by one. In particular, D is the final architecture.}
\label{tab:ablation}
\end{table}

\subsection{Final Results}

Final results are obtained by following the testing scheme described in the standard two-stream method~\cite{Simonyan14}. At first, 10 video snippets are randomly sampled and each of them contains 3 RGB images along with the corresponding 30 optical flow frames. We then enlarge the training datasets by cropping the frames and flipping them to avoid over-fitting. All data belonging to one snippet is fed to the network to produce an estimate, and the video-level prediction is made by averaging over the 10 snippets.

\begin{table}[h]
\begin{center}
\begin{tabular}{|l|c|c|}
\hline
Method & UCF101 & HMDB51 \\
\hline\hline
Slow Fusion CNN~\cite{Karpathy14} & 65.4\% & - \\
LRCN~\cite{donahue2015long} & 82.9\% & - \\
C3D~\cite{Tran15} & 85.2\% & - \\
Two-Stream (AlexNet)~\cite{Simonyan14} & 88.0\% & 59.4\% \\
Two-Stream + LSTM~\cite{Ng15} & 88.6\% & - \\
Two-Stream + Pooling~\cite{Ng15}  & 88.2\% & - \\
Transformation \cite{Wangxiaolong16} & 92.4\% & 62.0\% \\
Two-Stream (VGG-16)~\cite{Feichtenhofer16} & 90.6\% & 58.2\% \\
Two-Stream + Fusion~\cite{Feichtenhofer16} & 92.5\% & 65.4\% \\
TSN (BN-Inception)~\cite{Wang16} & 94.0\% & 68.5\% \\
\hline
Ours (VGG-16) & 93.2\% & 66.1\% \\
Ours (ResNet-50) & 93.8\% & 66.5\% \\
\textbf{Ours (BN-Inception)} & \textbf{94.6\%} & \textbf{68.9\%} \\
\hline
\end{tabular}
\end{center}
\caption{Performance comparison with the state-of-the-art.}
\label{tab:final}
\vspace{-5pt}
\end{table}

We compare the performance of our final architecture with the state-of-the-art in Table~\ref{tab:final}. Our best implementation based on BN-Inception improves the average accuracy by 0.6\% on UCF101 and 0.4\% on HMDB51. Someone may cast doubt on it and own this performance boost to the very deep models. To testify that our method is generally effective, we additionally use the same base architecture (VGG-16) as the previous two-stream CNNs architectures. Both based on VGG-16, our result (\textbf{93.2\%}) is still competitive to the original two-stream fusion \cite{Feichtenhofer16} (\textbf{92.5\%}). Again, both based on BN-Inception, our new result (\textbf{94.6\%}) is superior to the state-of-the-art \cite{Wang16} (\textbf{94.0\%}). This result also illustrate that our approaches are not any deep-network exclusive, but can be widely applied to many fancy CNN models. 

\begin{figure}[b]
\vspace{-10pt}
\begin{center}
\includegraphics[width=0.45\textwidth]{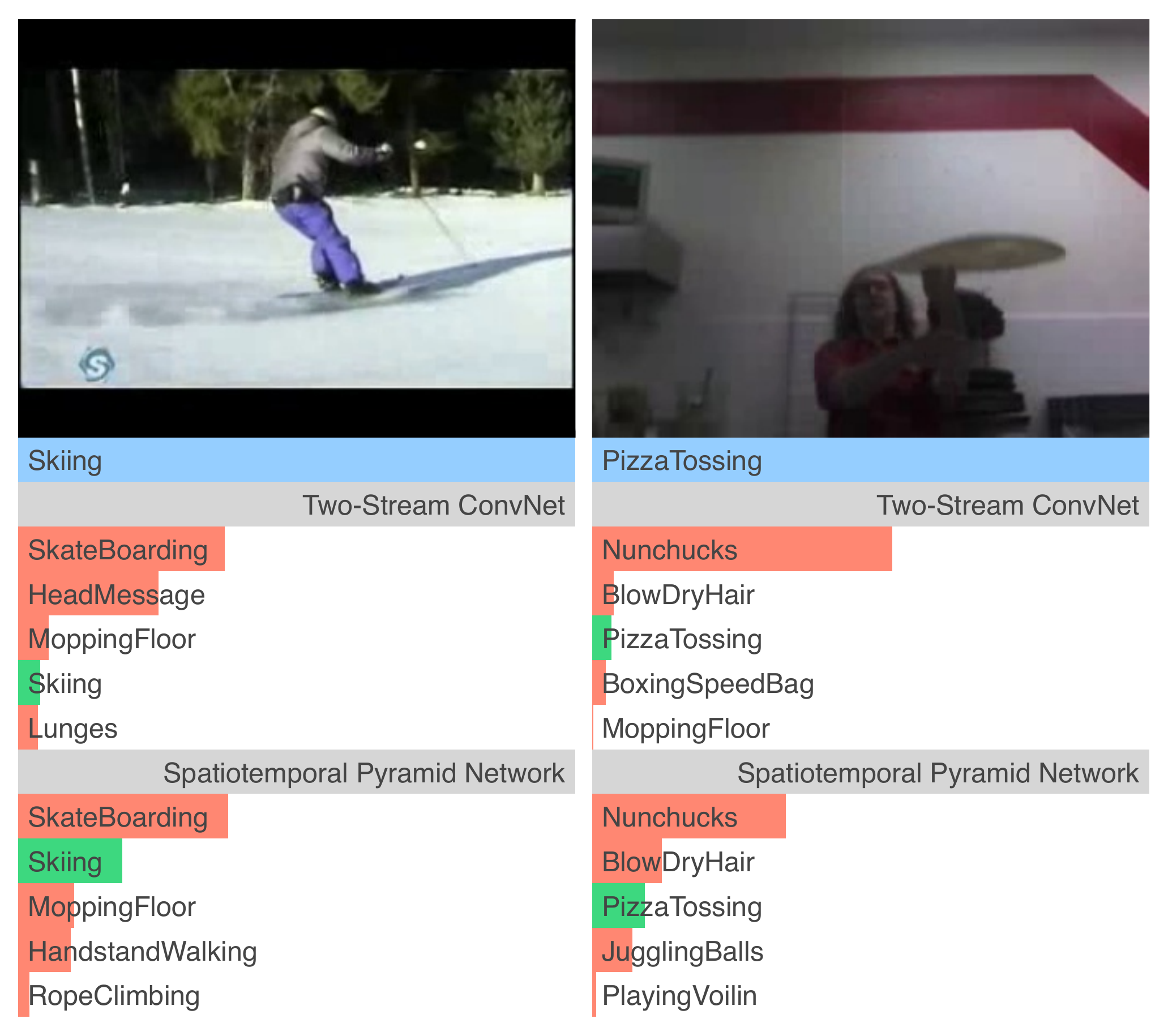}
\end{center}
   \caption{Examples of categories that are difficult to classify. Even for misclassification cases, the confidence of the correct category (green) has increased by our method.}
\label{fig:faults}
\end{figure}

\begin{figure*}
\begin{center}
\includegraphics[width=1.0\textwidth]{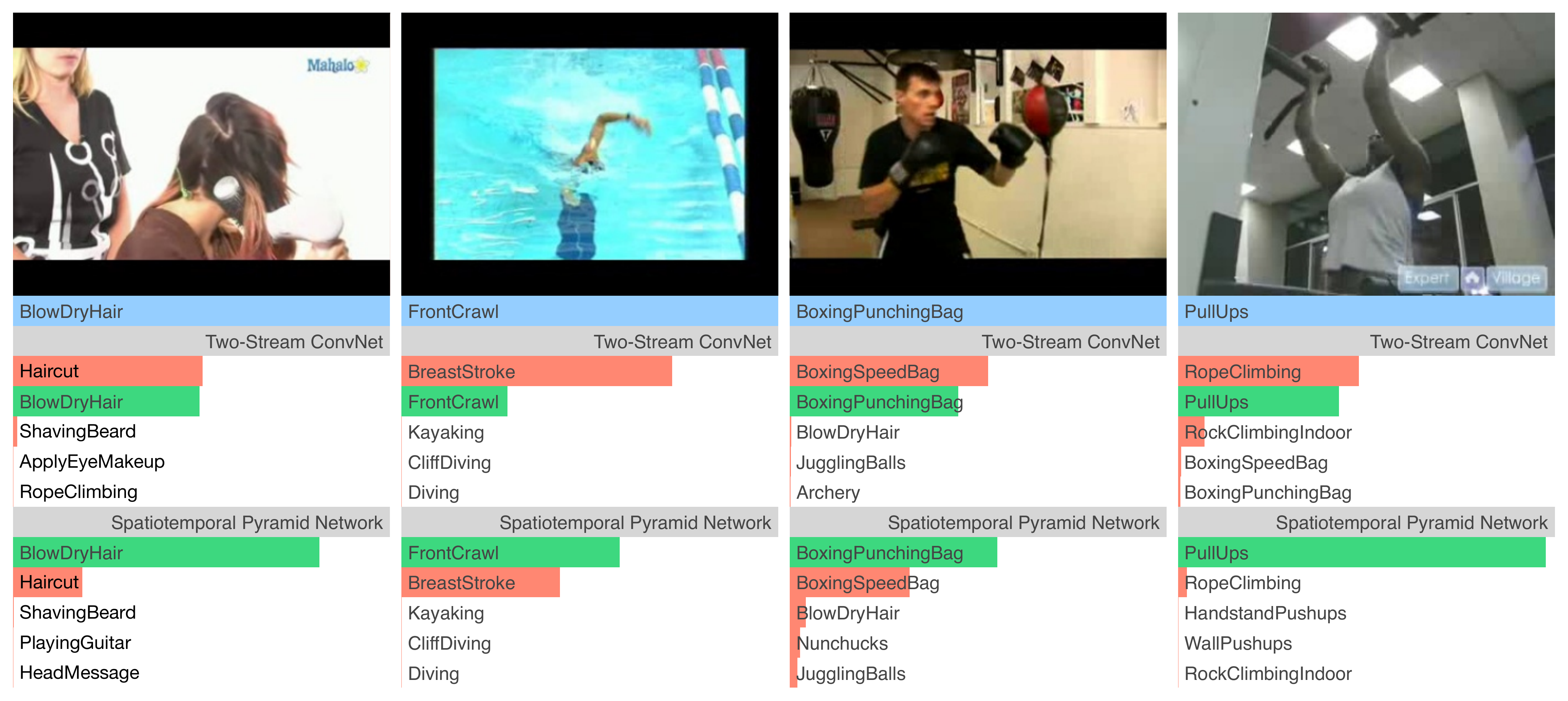}
\end{center}
\vspace{-10pt}
   \caption{A comparison of top-5 predictions between the baseline and our pyramid network on UCF-101. The blue bars denote the ground truth labels, the green bars indicate correct classifications and the red stand for incorrect cases. The length of each bar shows its confidence. With pyramid network, some errors can be eliminated by taking advantage of image attention (right two), while some other categories can be disambiguated by fusing long-term temporal features (left two).}
\label{fig:results}
\end{figure*}

\begin{figure*}[!phtb]
\begin{center}
\includegraphics[width=1.0\textwidth]{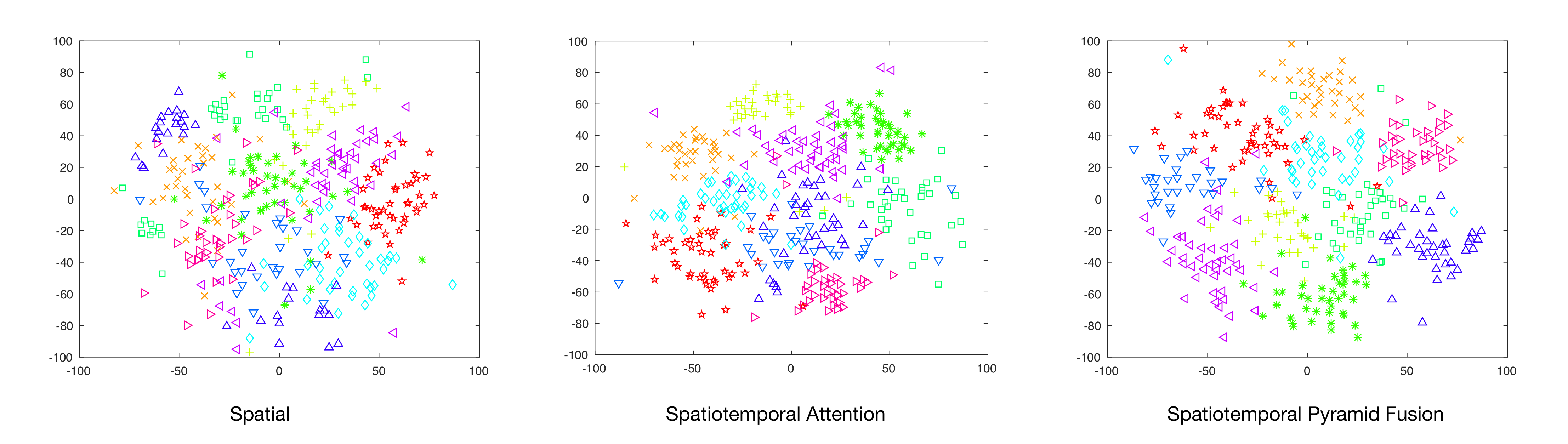}
\end{center}
   \caption{The t-SNE results of 10 classes randomly selected from UCF101. The left figure shows representations of an individual spatial network; the middle figure is obtained by adding the spatiotemporal attention method on it; and the right one denotes the results of our final spatiotemporal pyramid architecture. We use features of the classifier layer for all these cases.}
\label{fig:tSNE}
\vspace{-10pt}
\end{figure*}

Some representative examples of the classification results are shown in Figure~\ref{fig:results}. The first two subplots demonstrate the effectiveness of the spatiotemporal attention module. As mentioned above, the original two-stream network is easily fooled by common background. For instance, it regards FrontCrawl as BreastStroke, since the swimming pool appears to be a dominate feature. But in our model, these spatially ambiguous classes can be separated by exploiting motion information to extract attended regions of the activities. The last two subplots illustrate another strength of our pyramid network. Thanks to the multi-path temporal fusion, it produces more global features over longer video sequences and can easily differentiate actions that look similar in short-term snippets but may vary substantially in a long-term. Additionally, Figure~\ref{fig:faults} gives typical examples of categories that are difficult to classify. In the first case, the network sometimes regards Skiing as SkateBoarding. We can observe that the main difference of these two classes resides in the scene environment. A possible reason is that CNN is not robust to the color bias of the image background. There is no evidence that the attention truly ignores the background and harms the classification performance. On the contrary, it offers the fusion pyramid some useful and additional cues for accurate predictions. If observing Figure~\ref{fig:faults} carefully, one can find the confidence of the correct category (Skiing) has actually increased. This shows that one component may amend the error of others in the fusion pyramid. Moreover, the second example indicates that some categories, like PizzaTossing and Nunchucks, can only be disambiguated by taking advantage of a fine-grained recognition. That is to say, detecting the detailed objects in connection with the on-going actions is also important. The study towards this issue may reside in our future research.

Furthermore, we adopt t-SNE techniques to visualize feature vectors that are learned at different levels of the spatiotemporal pyramid, as shown in Figure~\ref{fig:tSNE}. We observe that the spatiotemporal attention can indeed improve the feature quality of the spatial stream, and the spatiotemporal compact bilinear fusion at the top of the pyramid can further increase the discriminative performance.

\section{Conclusions}
We propose a spatiotemporal pyramid network to combine the spatial and temporal features and make them reinforce each other. From the architecture perspective, our network is hierarchical, consisting of multiple fusion strategies at different abstraction levels. These fusion modules are trained as a whole to maximally complementing each other. A series of ablation studies validate the importance of each fusion technique. From the technical perspective, we introduce the spatiotemporal compact bilinear operator into video analysis tasks. This operator can learn element-wise interactions between the spatial and temporal features. We extensively show its benefit over other fusion methods, such as concatenation and element-wise sum. Our spatiotemporal pyramid network achieves the state-of-the-art performance on UCF101 and HMDB51.

\section*{Acknowledgments}
This work is supported by NSFC (61502265, 61325008), National Key R\&D Program of China (2016YFB1000701, 2015BAF32B01), Tsinghua National Laboratory (TNList) Key Projects, NSF through grants IIS-1526499 and CNS-1626432, and NSFC 61672313.


{\small
\bibliographystyle{ieee}
\bibliography{YWang}

\begin{thebibliography}{10}\itemsep=-1pt

\bibitem{Charikar2002}
M.~Charikar, K.~Chen, and M.~Farach-Colton.
\newblock {\em Finding Frequent Items in Data Streams}, pages 693--703.
\newblock Springer Berlin Heidelberg, 2002.

\bibitem{cheron2015p}
G.~Ch{\'e}ron, I.~Laptev, and C.~Schmid.
\newblock P-cnn: Pose-based cnn features for action recognition.
\newblock In {\em Proceedings of the IEEE International Conference on Computer
  Vision}, pages 3218--3226, 2015.

\bibitem{cimpoi2015deep}
M.~Cimpoi, S.~Maji, and A.~Vedaldi.
\newblock Deep filter banks for texture recognition and segmentation.
\newblock In {\em Proceedings of the IEEE Conference on Computer Vision and
  Pattern Recognition}, pages 3828--3836, 2015.

\bibitem{deng2009imagenet}
J.~Deng, W.~Dong, R.~Socher, L.-J. Li, K.~Li, and L.~Fei-Fei.
\newblock Imagenet: A large-scale hierarchical image database.
\newblock In {\em Computer Vision and Pattern Recognition, 2009. CVPR 2009.
  IEEE Conference on}, pages 248--255. IEEE, 2009.

\bibitem{donahue2015long}
J.~Donahue, L.~Anne~Hendricks, S.~Guadarrama, M.~Rohrbach, S.~Venugopalan,
  K.~Saenko, and T.~Darrell.
\newblock Long-term recurrent convolutional networks for visual recognition and
  description.
\newblock In {\em Proceedings of the IEEE Conference on Computer Vision and
  Pattern Recognition}, pages 2625--2634, 2015.

\bibitem{Feichtenhofer16}
C.~Feichtenhofer, A.~Pinz, and A.~Zisserman.
\newblock Convolutional two-stream network fusion for video action recognition.
\newblock In {\em IEEE Conference on Computer Vision and Pattern Recognition},
  2016.

\bibitem{gao2016compact}
Y.~Gao, O.~Beijbom, N.~Zhang, and T.~Darrell.
\newblock Compact bilinear pooling.
\newblock In {\em IEEE Conference on Computer Vision and Pattern Recognition},
  2016.

\bibitem{He2015Deep}
K.~He, X.~Zhang, S.~Ren, and J.~Sun.
\newblock Deep residual learning for image recognition.
\newblock In {\em IEEE Conference on Computer Vision and Pattern Recognition},
  2016.

\bibitem{ioffe2015batch}
S.~Ioffe and C.~Szegedy.
\newblock Batch normalization: Accelerating deep network training by reducing
  internal covariate shift.
\newblock In {\em Proceedings of the International Conference on Machine
  Learning (ICML)}, 2015.

\bibitem{Ji13}
S.~Ji, W.~Xu, M.~Yang, and K.~Yu.
\newblock 3d convolutional neural networks for human action recognition.
\newblock {\em IEEE transactions on pattern analysis and machine intelligence},
  35(1):221--231, 2013.

\bibitem{Jia14}
Y.~Jia, E.~Shelhamer, J.~Donahue, S.~Karayev, J.~Long, R.~Girshick,
  S.~Guadarrama, and T.~Darrell.
\newblock Caffe: Convolutional architecture for fast feature embedding.
\newblock In {\em Proceedings of the 22nd ACM international conference on
  Multimedia}, pages 675--678. ACM, 2014.

\bibitem{Karpathy14}
A.~Karpathy, G.~Toderici, S.~Shetty, T.~Leung, R.~Sukthankar, and L.~Fei-Fei.
\newblock Large-scale video classification with convolutional neural networks.
\newblock In {\em Proceedings of the IEEE conference on Computer Vision and
  Pattern Recognition}, pages 1725--1732, 2014.

\bibitem{krizhevsky2012imagenet}
A.~Krizhevsky, I.~Sutskever, and G.~E. Hinton.
\newblock Imagenet classification with deep convolutional neural networks.
\newblock In {\em Advances in neural information processing systems}, pages
  1097--1105, 2012.

\bibitem{Kuehne12}
H.~Kuehne, H.~Jhuang, E.~Garrote, T.~Poggio, and T.~Serre.
\newblock Hmdb: a large video database for human motion recognition.
\newblock In {\em 2011 International Conference on Computer Vision}, pages
  2556--2563. IEEE, 2011.

\bibitem{laptev2008learning}
I.~Laptev, M.~Marszalek, C.~Schmid, and B.~Rozenfeld.
\newblock Learning realistic human actions from movies.
\newblock In {\em Computer Vision and Pattern Recognition, 2008. CVPR 2008.
  IEEE Conference on}, pages 1--8. IEEE, 2008.

\bibitem{lecun1998gradient}
Y.~LeCun, L.~Bottou, Y.~Bengio, and P.~Haffner.
\newblock Gradient-based learning applied to document recognition.
\newblock {\em Proceedings of the IEEE}, 86(11):2278--2324, 1998.

\bibitem{lin2015bilinear}
T.-Y. Lin, A.~RoyChowdhury, and S.~Maji.
\newblock Bilinear cnn models for fine-grained visual recognition.
\newblock In {\em Proceedings of the IEEE International Conference on Computer
  Vision}, pages 1449--1457, 2015.

\bibitem{lu2016hierarchical}
J.~Lu, J.~Yang, D.~Batra, and D.~Parikh.
\newblock Hierarchical question-image co-attention for visual question
  answering.
\newblock In {\em Advances in Neural Information Processing Systems (NIPS)},
  2016.

\bibitem{perronnin2010improving}
F.~Perronnin, J.~S{\'a}nchez, and T.~Mensink.
\newblock Improving the fisher kernel for large-scale image classification.
\newblock In {\em European conference on computer vision}, pages 143--156.
  Springer, 2010.

\bibitem{Pham2013}
N.~Pham and R.~Pagh.
\newblock Fast and scalable polynomial kernels via explicit feature maps.
\newblock In {\em Proceedings of the 19th ACM SIGKDD International Conference
  on Knowledge Discovery and Data Mining}, KDD '13, pages 239--247, New York,
  NY, USA, 2013. ACM.

\bibitem{Sharma16}
S.~Sharma, R.~Kiros, and R.~Salakhutdinov.
\newblock Action recognition using visual attention.
\newblock In {\em ICLR}, 2016.

\bibitem{Simonyan14}
K.~Simonyan and A.~Zisserman.
\newblock Two-stream convolutional networks for action recognition in videos.
\newblock In {\em Advances in Neural Information Processing Systems}, pages
  568--576, 2014.

\bibitem{simonyan2015very}
K.~Simonyan and A.~Zisserman.
\newblock Very deep convolutional networks for large-scale image recognition.
\newblock In {\em Proceedings of the International Conference on Learning
  Representations (ICLR)}, 2015.

\bibitem{Soomro12}
K.~Soomro, A.~R. Zamir, and M.~Shah.
\newblock Ucf101: A dataset of 101 human actions classes from videos in the
  wild.
\newblock {\em arXiv preprint arXiv:1212.0402}, 2012.

\bibitem{srivastava2015unsupervised}
N.~Srivastava, E.~Mansimov, and R.~Salakhutdinov.
\newblock Unsupervised learning of video representations using lstms.
\newblock {\em CoRR, abs/1502.04681}, 2, 2015.

\bibitem{sun2015human}
L.~Sun, K.~Jia, D.-Y. Yeung, and B.~E. Shi.
\newblock Human action recognition using factorized spatio-temporal
  convolutional networks.
\newblock In {\em Proceedings of the IEEE International Conference on Computer
  Vision}, pages 4597--4605, 2015.

\bibitem{szegedy2015going}
C.~Szegedy, W.~Liu, Y.~Jia, P.~Sermanet, S.~Reed, D.~Anguelov, D.~Erhan,
  V.~Vanhoucke, and A.~Rabinovich.
\newblock Going deeper with convolutions.
\newblock In {\em Proceedings of the IEEE Conference on Computer Vision and
  Pattern Recognition}, pages 1--9, 2015.

\bibitem{Tran15}
D.~Tran, L.~Bourdev, R.~Fergus, L.~Torresani, and M.~Paluri.
\newblock Learning spatiotemporal features with 3d convolutional networks.
\newblock In {\em 2015 IEEE International Conference on Computer Vision
  (ICCV)}, pages 4489--4497. IEEE, 2015.

\bibitem{venugopalan2015sequence}
S.~Venugopalan, M.~Rohrbach, J.~Donahue, R.~Mooney, T.~Darrell, and K.~Saenko.
\newblock Sequence to sequence-video to text.
\newblock In {\em Proceedings of the IEEE International Conference on Computer
  Vision}, pages 4534--4542, 2015.

\bibitem{wang2013action}
H.~Wang and C.~Schmid.
\newblock Action recognition with improved trajectories.
\newblock In {\em Proceedings of the IEEE International Conference on Computer
  Vision}, pages 3551--3558, 2013.

\bibitem{wang2015action}
L.~Wang, Y.~Qiao, and X.~Tang.
\newblock Action recognition with trajectory-pooled deep-convolutional
  descriptors.
\newblock In {\em Proceedings of the IEEE Conference on Computer Vision and
  Pattern Recognition}, pages 4305--4314, 2015.

\bibitem{Wang16}
L.~Wang, Y.~Xiong, and et~al.
\newblock Temporal segment networks: Towards good practices for deep action
  recognition.
\newblock In {\em ECCV}, 2016.

\bibitem{wang2015towards}
L.~Wang, Y.~Xiong, Z.~Wang, and Y.~Qiao.
\newblock Towards good practices for very deep two-stream convnets.
\newblock {\em arXiv preprint arXiv:1507.02159}, 2015.

\bibitem{Wangxiaolong16}
X.~Wang, A.~Farhadi, and A.~Gupta.
\newblock Actions ~ transformations.
\newblock In {\em Proceedings of the IEEE conference on Computer Vision and
  Pattern Recognition}, pages 2658--2667, 2016.

\bibitem{weinzaepfel2015learning}
P.~Weinzaepfel, Z.~Harchaoui, and C.~Schmid.
\newblock Learning to track for spatio-temporal action localization.
\newblock In {\em Proceedings of the IEEE International Conference on Computer
  Vision}, pages 3164--3172, 2015.

\bibitem{xu2015show}
K.~Xu, J.~Ba, R.~Kiros, K.~Cho, A.~Courville, R.~Salakhutdinov, R.~S. Zemel,
  and Y.~Bengio.
\newblock Show, attend and tell: Neural image caption generation with visual
  attention.
\newblock In {\em Proceedings of the International Conference on Machine
  Learning (ICML)}, 2015.

\bibitem{yang2015stacked}
Z.~Yang, X.~He, J.~Gao, L.~Deng, and A.~Smola.
\newblock Stacked attention networks for image question answering.
\newblock {\em arXiv preprint arXiv:1511.02274}, 2015.

\bibitem{Ng15}
J.~Yue-Hei~Ng, M.~Hausknecht, S.~Vijayanarasimhan, O.~Vinyals, R.~Monga, and
  G.~Toderici.
\newblock Beyond short snippets: Deep networks for video classification.
\newblock In {\em Proceedings of the IEEE Conference on Computer Vision and
  Pattern Recognition}, pages 4694--4702, 2015.

\bibitem{zhang2016real}
B.~Zhang, L.~Wang, Z.~Wang, Y.~Qiao, and H.~Wang.
\newblock Real-time action recognition with enhanced motion vector cnns.
\newblock In {\em IEEE Conference on Computer Vision and Pattern Recognition},
  2016.

\end{thebibliography}
}

\end{document}